# Detection of Pavement Cracks by Deep Learning Models of Transformer and UNet

Yu Zhang and Lin Zhang

*Abstract*—Fracture is one of the main failure modes of engineering structures such as buildings and roads. Effective detection of surface cracks is significant for damage evaluation and structure maintenance. In recent years, the emergence and development of deep learning techniques have shown great potential to facilitate surface crack detection. Currently, most reported tasks were performed by a convolutional neural network (CNN), while the limitation of CNN may be improved by the transformer architecture introduced recently. In this study, we investigated nine promising models to evaluate their performance in pavement surface crack detection by model accuracy, computational complexity, and model stability. We created 711 images of 224 by 224 pixels with crack labels, selected an optimal loss function, compared the evaluation metrics of the validation dataset and test dataset, analyzed the data details, and checked the segmentation outcomes of each model. We find that transformer-based models generally are easier to converge during the training process and have higher accuracy, but usually exhibit more memory consumption and low processing efficiency. Among nine models, SwinUNet outperforms the other two transformers and shows the highest accuracy among nine models. The results should shed light on surface crack detection by various deep-learning models and provide a guideline for future applications in this field.

## I. INTRODUCTION

IN concrete roads, pavement surface cracks are significant to road safety. The routine maintenance and life prediction of the roads is based on the damage evaluation, which mainly depends on how accurate to detect the pavement surface cracks. Usually, manual inspection methods are applied to detect pavement cracks. Those methods are highly labor-intensive and time-consuming, and traffic control or blocking is necessary for the safety of the inspectors. This method relies heavily on the subjective judgment of the testers, the working experience, and the physical condition of the personnel, which may have an impact on the testing results [1]. With the development of computer vision (CV), image segmentation, a computer vision technology, has been applied to pavement crack detection. Initially, this technique is based on the global threshold and edge features of the image to distinguish cracks from the background in the images. However, when the road noise, shadow, oil pollution, and other disturbances are randomly incorporated, or the contrast between crack pixels and road background is low, this technology lacks a good characterization in crack detection [2].

Deep learning (DL) has achieved remarkable progress in computer vision (CV) tasks in the past decade, which promotes crack detection development. The convolutional neural network (CNN) model is a representative, which has been applied to crack detection. Most efforts have been made how to create more effective deep-learning models [3]. There are two categories of crack identification models, namely patch-based methods and pixel-based methods. The former is based on the classification task in CV to judge whether there is a crack in the picture or to determine the location of the crack [2]. Cha and Choi built a classifier using CNN on the crack dataset of size 256×256 pixels, combined with sliding window techniques to obtain coarse localization of cracks in images arbitrarily larger than this pixel size [4]. Cha et al. proposed to apply the Fast Region-CNN model to detect concrete cracks, steel corrosion, and bolt corrosion, achieving the purpose of real-time synchronous detection [5]. Chang and Zhang et al. improved the original YOLOv3 model and introduced a new transfer learning method to enhance further the efficiency of simultaneous crack detection [6]. They achieved higher accuracy under the same Intersection over Union (IoU) metric. All these studies use classification neural networks or generate bounding boxes for the object detection of cracks. Despite the impressive progress, these methods can not provide pix-level details of the cracks, let alone the precise dimensions of the crack. To this end, historical information of cracks can not be provided for damage evolution and life analysis.

Instead, crack pixel-level detection is developed by deep learning-based image segmentation techniques and can provide pixel-level details [2]. Yang et al. proposed a new network structure, the Feature Pyramid and Hierarchical Boosting Network (FPHBN). This network enriches low-level features by integrating semantic information from high-level layers in a pyramidal fashion, and it turns out that the network is superior to edge detection and semantic segmentation methods [7]. The deep convolutional neural network (DCNN) trained by Zhang et al. based on the transfer learning strategy realizes the region pre-classification of road images so that noisy background regions can be filtered out. Along with the threshold segmentation method, Zhang et al. successfully divided the crack region to the pixel level and significantly improved the performance over the previous machine learning algorithm [8].

.

This work was supported in part by the Shandong Provincial Natural Science Foundation under Grant ZR2021MA045. The scientific calculations have been performed on the HPC Cloud Platform of Shandong University. (*Corresponding author: Lin Zhang*).

Yu Zhang and Lin Zhang are with the Department of Engineering Mechanics, the School of Civil Engineering, Shandong University, Jinan 250061, China (e-mail: linzhang1629@email.sdu.edu.cn).

Color versions of one or more of the figures in this article are available online at http://ieeexplore.ieee.org

Although DCNN algorithms are relatively mature, they do have certain limitations. For example, they suffer feature loss as the number of convolutions and pooling increases.

Recently, an end-to-end method to predict pixel-level crack segmentation is presented by building an encoder-decoder structure [2]. Typical algorithms include a fully convolutional network (FCN) [3, 9, 10], DeepCrack [11], DeepLabv3+ [12, 13], and UNet [1, 14-16]. UNet stemmed from FCN and was first proposed by Ronneberger et al. for biomedical image segmentation applications [14]. After that, Liu et al. used the UNet method to detect concrete cracks and compared it with DCNN-based methods. They found that UNet is more robust, effective, and accurate in detection than DCNN [15]. Although the original UNet model can be successfully transferred to the crack detection task, it may not be suitable for detecting cracks in many complex situations. Inspired by UNet, Zhang et al. proposed an improved pixel-level surface crack detection model called CrackUnet. The CrackUnet model outperformed other methods when comparing the detection accuracy, training time, and data set information [17]. By mixing UNet with ResNet [18], MobileNets [19], and other efficient network models, Dais and Liu comprehensively compared the performance of various hybrid networks in masonry crack detection and concrete crack detection, respectively [1, 16]. Except for minor modifications in the architecture of UNet itself, theoretically, various end-to-end networks for image segmentation tasks, such as UNet ++ [20, 21], Deep Residual UNet (ResUNet) [22], U2Net [23], ResUNet++ [24], Attention UNet (AttnUNet) [25], etc., can be used for crack detection through transfer learning. As the core of those methods, CNN is the main algorithm in deep learning and greatly promotes the development of deep learning.

Recently, a new information processing architecture called transformer gradually surpassed CNN in the latest computer vision tasks. Transformer was originally proposed as a model structure for natural language processing (NLP) through self-attention (SA), as shown in **Fig. 1.** Transformer completely abandoned the traditional CNN and Recurrent Neural Network (RNN) model algorithms and achieved a significant improvement in the NLP field only through SA and feedforward neural networks [25]. When the transformer model was widely used in NLP, Google's research team successfully applied the transformer to computer vision. By dividing the image into 16 × 16 patches and importing them into SA, the team proposed a computer vision model that completely abandoned CNN, called the Vision Transformer (ViT). And they found that the ViT model can provide better results and requires fewer computing resources compared to CNN when the scale of the data set is large [26]. However, for images with high resolution and many pixels, ViT based on global self-attention cannot guarantee a small amount of calculation. To address such issues, Liu et al. presented a new vision Transformer, Swin Transformer, and uses a hierarchical transformer whose representation is computed with shifted windows. Liu et al. also pointed out that this model has not only good performance on classification tasks but also wide compatibility in downstream tasks of CV, such as detection and segmentation [27]. As a result, the model has taken the top spot

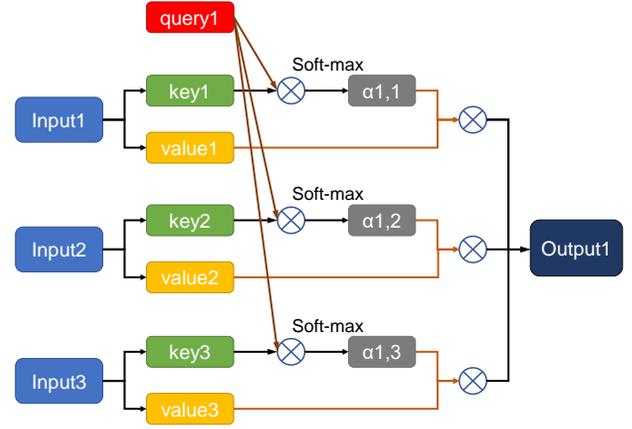

**Fig. 1.** The detailed structure of self-attention (SA).

in various CV tasks since its release.

Considering that UNet usually exhibits limitations in explicitly modeling long-term dependencies due to the inherent locality of convolution operations, Chen et al. proposed the TransUNet model [26, 27]. The hybrid CNN-Transformer architecture adopted by TransUNet combines the advantages of UNet in low-level information extraction and the transformer's nature of the global context modeling, making TransUNet a powerful alternative to medical image segmentation. After TransUNet successfully fused UNet and transformer, Cao et al. proposed SwinUNet. Unlike the former hybrid structure, SwinUNet adopts a complete SA algorithm for decoding and encoding and uses Swin Transformer with shifted windows to extract contextual features [28]. On the medical multi-organ image segmentation task, SwinUNet outperforms TransUNet and other CNN methods. Recently, Wang et al. summarized the defects of SA in Transformer. First, SA requires large-scale pre-training. Second, SA has quadratic computational complexity. Last, SA has the limitation of ignoring the correlation between samples [29]. Based on this, Wang et al. redesigned SA, introduced external attention [30], and proposed Mixed Transformer UNet (MTUNet). It combines the well-designed Local-Global Gaussian-Weighted Self-Attention (LGG-SA) and External Attention (EA) modules, making another progress in medical image segmentation.

Although transformers have achieved excellent performance in medical image segmentation, few researchers have applied them to pavement crack detection. In order to find a more accurate crack detection model, we demonstrate the feasibility of the transformer in crack detection and evaluate it by the model accuracy, computational complexity, and model stability. A crack segmentation dataset was created with 711 pavement crack images of a size of 224 × 224 pixels. The dataset considered the applicability in actual situations, such as disturbance by oil, vegetation, and shadows. In this paper, transformer-based image segmentation models, including TransUNet, SwinUNet, and MTUNet, are applied to pavement crack detection. Their performances are compared with the previous CNN crack detection model, and the loss function and accuracy metrics in the model training and validation process are comprehensively analyzed. On the other hand, The performance of transformer models on crack segmentation is evaluated by model operation speed, memory, and model accuracy. The results should shed light on surface crack detection by various deep-learning models and provide a guideline for future applications in this field.

## II. METHOD

### A. TransUNet architecture

This section describes the detailed architecture of the TransUNet network, as shown in **Fig. 2**.

*1) Extract features using CNN:* In the encoder part (see **Fig. 2a**), TranUNet uses CNN first to extract low-level features on raw images instead of the pure transformer for encoding because it considers that ViT [31] has a limited ability to extract low-level features of images. The ViT is based on global self-attention and requires a large amount of calculation, so the effect of ViT may not be obvious.

*2) Patch embedding and position embedding:* In the original ViT, we reshape the input map $x \in R^{H \times W \times C}$ into a series of 2D patches $\{x_p^i \in R^{P^2 \cdot C} | i = 1, .., N\}$, where the spatial resolution is $H \times W$, the number of channels is C, and the 2D patch is of size $P \times P$. The number of patches that can be easily obtained is $N = \frac{HW}{P^2}$. Subsequently, the trainable linear projections $E \in R^{(P^2 \cdot C) \times D}$ are applied to the vectorized patch to map the vectorized patch into a latent D-dimensional embedding space. Meanwhile, we add the position embedding block $E_{pos} \in R^{N \times D}$ to the patch embeddings for the position encoding of patch spatial information, as shown by

$$Z_0 = [x_p^1 E; x_p^2 E; ...; x_p^N E] + E_{pos}, \quad (1)$$

It is worth noting that for segmentation, TranUNet extracts 1 × 1 patches from the feature map obtained after CNN through patch embedding. A coded feature representation is then generated through mapping and positional encoding.

*3) Transformer layer:* Like ViT, the transformer layer part of the encoder consists of Multi-head Self-Attention (MSA) and Multi-Layer Perceptron (MLP) (see **Fig. 2b**). Therefore, the transformer layer process of the *l*-th layer is shown as follows:

$$z_l' = MSA(LN(z_{l-1})) + z_{l-1}, \quad (2)$$

$$z_l = MLP(LN(z_l')) + z_l', \quad (3)$$

where LN(~) is expressed as a layer normalization operator, $z_l'$ is an intermediate variable.

*4) Upsample:* TranUNet designs a continuous upsampling decoder, which first reshapes the hidden feature obtained through the transformer layer into a tensor form that can be continuously upsampled. We can finally determine the semantic information by continuous upsampling and skip connection with the feature map obtained by CNN (see **Fig. 2a**).

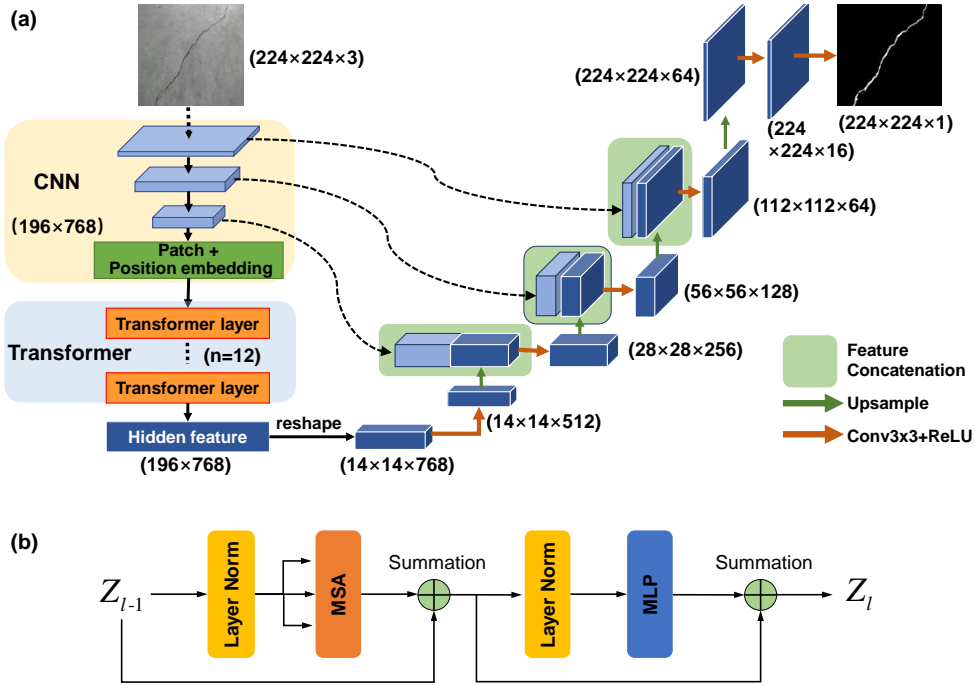

**Fig. 2.** The detailed architecture of the TransUNet with (a) the overall architecture and (b) the transformer layer.

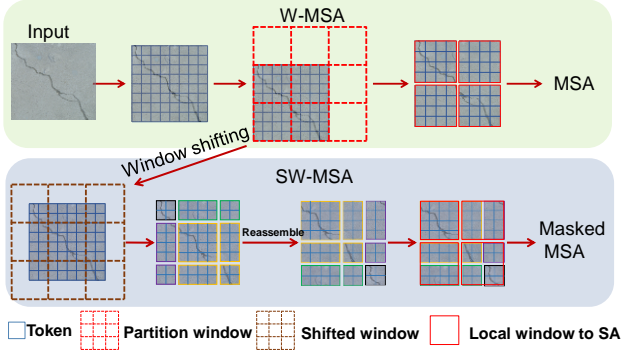

**Fig. 3.** Moving Window Methods and Window Partitioning Schemes in W-MSA and SW-MSA.

*B. SwinUNet architecture*

This section describes SwinUNet, a network that adopts the idea of a Swin transformer [32] and uses pure transformers for image segmentation.

*1) W-MSA and SW-MSA:* Compared with ViT, the most critical improvement of the Swin transformer is to use Windows Multi-head Self-Attention (W-MSA) and Shifted Windows Multi-head Self-Attention (SW-MSA) to replace the global self-attention in ViT.

As shown in **Fig. 3**, the input is first divided into individual patches by patch division similar to ViT, and these patches can be called tokens. Second, the patches are organized into serval partial windows by the red dotted partition window. The self-attention relationship between one token and all other tokens in each local window is computed to reduce the amount of computation. These two steps are called W-MSA. This treatment can only focus on local information and cannot extract global features. To solve this problem, researchers proposed the SW-MSA method by shifting the partition window in an oblique direction after performing W-MSA. Obviously, the resulting blocks divided by the shifted window are not in the same shape. The blocks are reassembled to ensure the consistency of the size of the local window. The reassembly results in an issue that tokens next to each other in a partial window may have large differences because they are not adjacent initially. SW-MSA applies a masked MSA to limit the self-attention computation to each local window by fully considering the input position information and reducing the contact between non-adjacent tokens.

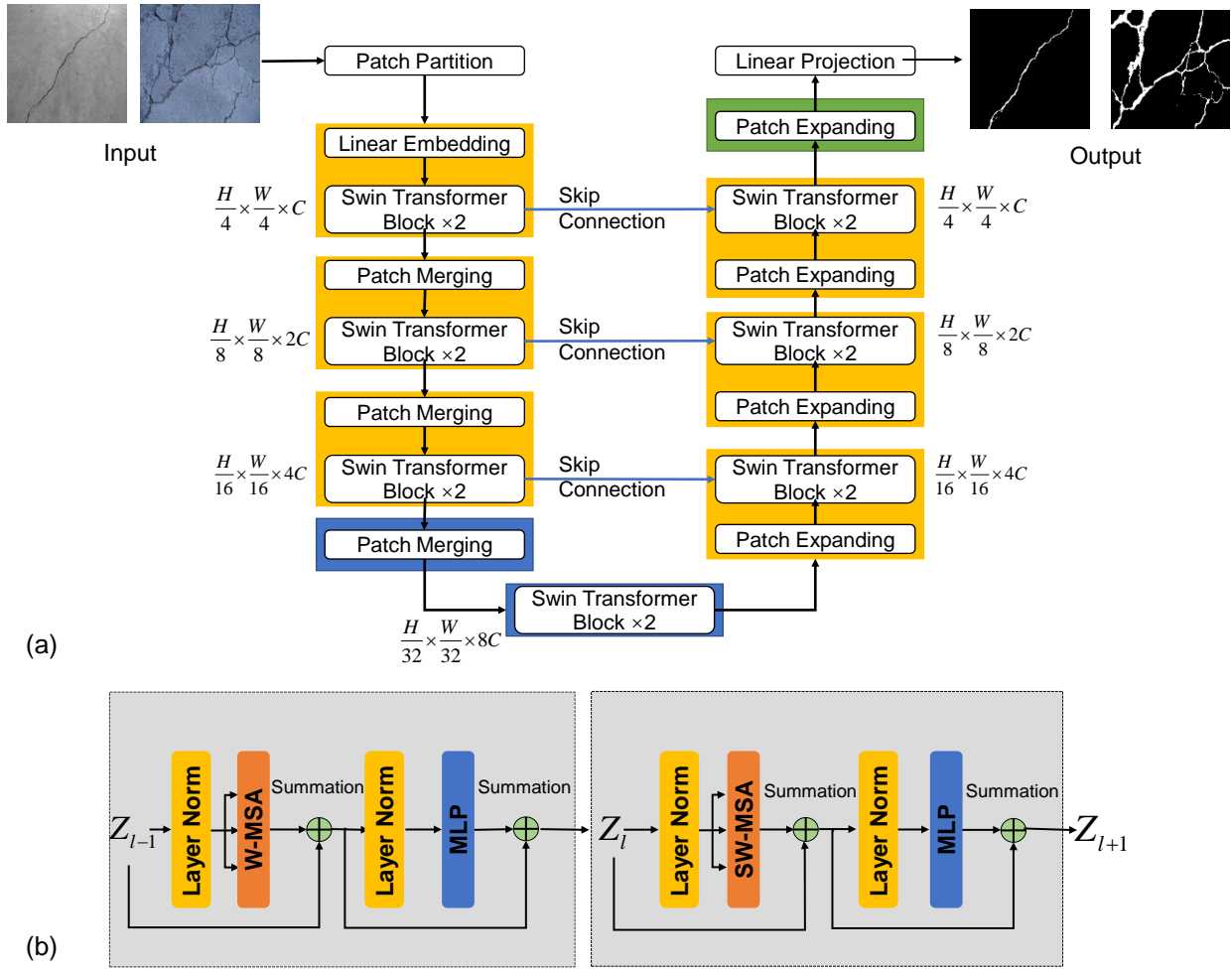

**Fig. 4.** (a) The overall architecture of SwinUNet, (b) Two consecutive transformer blocks.

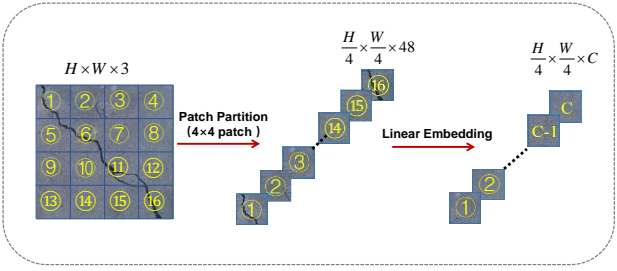

**Fig. 5**. Patch Partition and Linear Embedding.

*2) The overall structure of SwinUNet:* The overall architecture of SwinUNet, shown in **Fig. 4,** has an encoder-decoder architecture similar to UNet. In the encoder part, the input $x \in R^{H \times W \times 3}$ is first divided into 4 × 4 patches through Patch Partition(see **Fig. 5**) and then converted into sequence embeddings $z_0 \in R^{\frac{H}{4} \times \frac{W}{4} \times 48}$ Furthermore, a linear embedding layer is applied to project the feature dimension to an arbitrary dimension, denoted as C. The generated tokenized inputs go through several Swin transformer blocks (see **Fig. 4b**) and patch merging layers to generate feature representations at different levels. Here, the patch merging layer is used for down-sampling. Two consecutive transformer blocks can be written as:

$$z'_l = W\text{-}MSA(LN(z_{l-1})) + z_{l-1} \quad (4)$$

$$z_l = MLP(LN(z'_l)) + z'_l \quad (5)$$

$$z'_{l+1} = SW\text{-}MSA(LN(z_l)) + z_l \quad (6)$$

$$z_{l+1} = MLP(LN(z'_{l+1})) + z'_{l+1} \quad (7)$$

In the decoder part, the same transformer blocks as in the encoder are used for feature learning and up-sampling through patch expanding. In this part, the patch expansion layer can reconstruct the resolution of the feature map to twice the original and correspondingly reduce the feature dimension to half. To avoid losing semantic information in the feature extraction process, skip connections inspired by UNet are applied before the decoder and encoder. As a result, the deep and shallow features of the same dimension are fused.

*C. MTUNet architecture*

Similar to UNet and SwinUNet, MTUNet is also a u-shaped architecture with skip connections between the decoder and the encoder (see **Fig. 6 a**). The difference is that MTUNet uses a mixture of CNN and Mixed Transformer Module (MTM) (see **Fig. 6 b**) for feature learning. In addition, up-sampling and down-sampling in MTUNet take 3 × 3 convolution or deconvolution with a stride of 2, respectively.

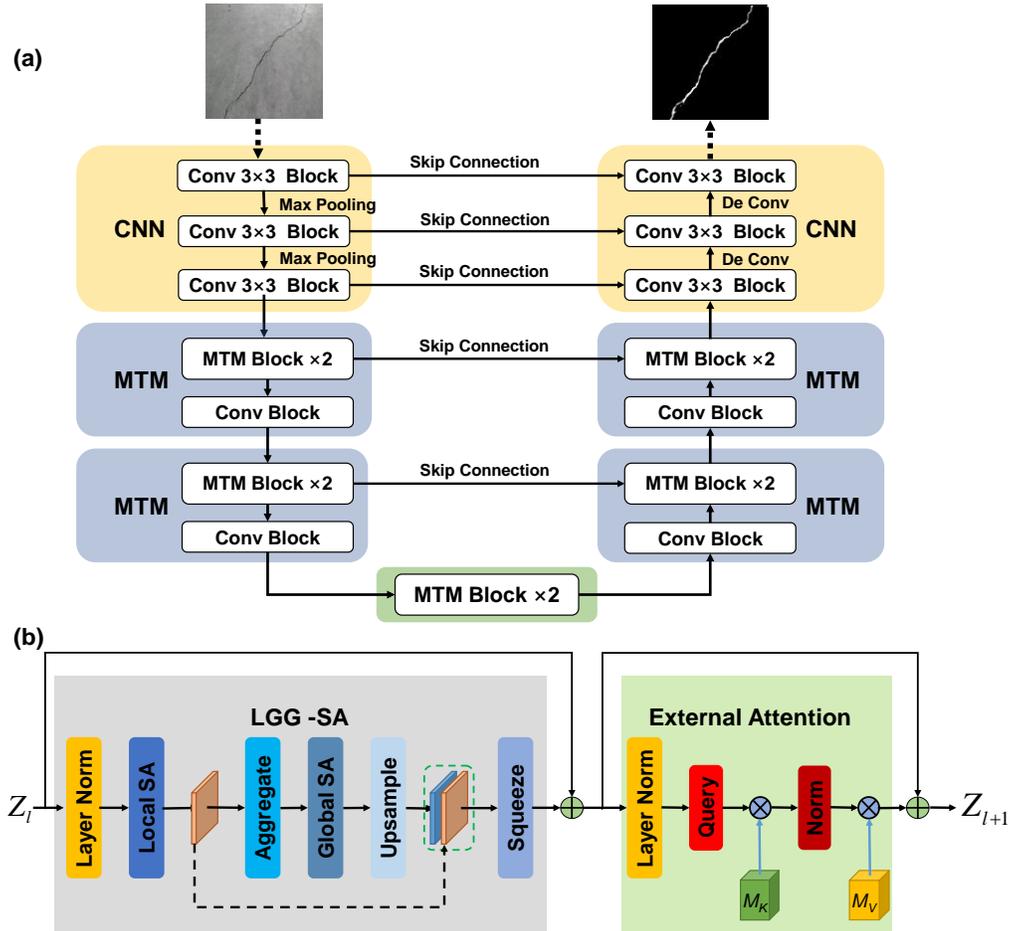

**Fig. 6.** (a) The overall architecture of MTUNet, (b) Two consecutive MTM blocks.

*1)Local-Global Gaussian-Weighted Self-Attention (LGG-SA):* It is reasonable that correlations between nearby regions tend to be more critical than those far away when SA computation is applied to feature maps. Therefore, LGG-SA focuses more on nearby local regions through a global-local strategy and a Gaussian mask. This is quite different from how the global SA is calculated in ViT.

As shown in **Fig. 7**, LGG-SA is similar to W-MSA in SwinUNet and divides the input feature map $x \in R^{H \times W \times C}$ into several $p \times p$ local windows, and $p$ is set to 4 in our experiment. We only calculate the SA between tokens in the same window and output the feature learning result $z_{local} \in R^{H \times W \times C}$ of the same dimension as *x*. The above process is called Local Self-Attention (LSA). Subsequently, the aggregation function of Dynamic Convolution (LDConv) [33] is used to down-sample the window generated by SA and reorganize it into a feature map, which is the input of Global Self-Attention (GSA).

In Global Self-Attention, Gaussian-Weighted Axial Attention (GWAA) is used to replace the traditional SA. In detail, $D_{i,j}$ is defined as the Euclidean distance matrix between the query $Q_{i,j}$ and the corresponding key $K_{i,j}$ as well as the value $V_{i,j}$, where *i* and *j* represent the row and column of the corresponding token. In addition, when calculating the similarity between $Q_{i,j}$ and $K_{i,j}$, it is only necessary to calculate the SA in the row and column directions where the corresponding token is located, which is called Axial Attention [34]. We assume that this similarity is $S(Q_{i,j}, K_{i,j})$ and the Gaussian weight is $e^{-\frac{D_{i,j}^2}{2\sigma^2}}$ so that the final output can be written as :

$$z_{i,j} = [e^{-\frac{D_{i,j}^2}{2\sigma^2}} \cdot Softmax(S(Q_{i,j}, K_{i,j}))] \cdot V_{i,j} \quad (8)$$

Considering the learnability of Gaussian weights, (8) can be further written as:

$$z_{i,j} = Softmax(-\frac{D_{i,j}^2}{2\sigma^2} + S(Q_{i,j}, K_{i,j})) \cdot V_{i,j} \quad (9)$$

Finally, the output result, $z_{local}$, of LSA and the up-sampled GSA output are fused through feature concatenation to realize global and local information exchange. The above process can be written as:

$$z_{local} = LSA(x) \quad (10)$$

$$z_{global} = GSA(LDConv(z_{local})) \quad (11)$$

$$z = Concat(z_{local}, Upsample(z_{global})) \quad (12)$$

*2)External Attention (EA):* An obvious disadvantage of SA is that there is a certain amount of computational redundancy, and it only performs feature learning through Q, K, and V based on its own linear transformation in the same sample, ignoring the correlation between different samples. EA [30] introduces two external memory units MK and MV to represent the features shared by all input samples and uses two cascaded linear layers and normalization layers to realize the learning of input features (see **Fig. 6.b**).

*3)The overall structure of MTUNet:* As shown in **Fig. 6.a**, MTUNet still uses a symmetric decoder-encoder architecture, and skip connections are made between them to transfer the semantic information. At the same time, due to the lack of prior knowledge of the problem, the transformer often requires large-scale pre-training, so MTUNet uses CNN as a shallow feature extractor. For deep features, MTM is used to extract semantic information, and the dimension of feature maps is reduced or increased through convolution or deconvolution. It is worth noting that two consecutive MTM blocks include LGG-SA and EA, which are shown in **Fig. 6.b**.

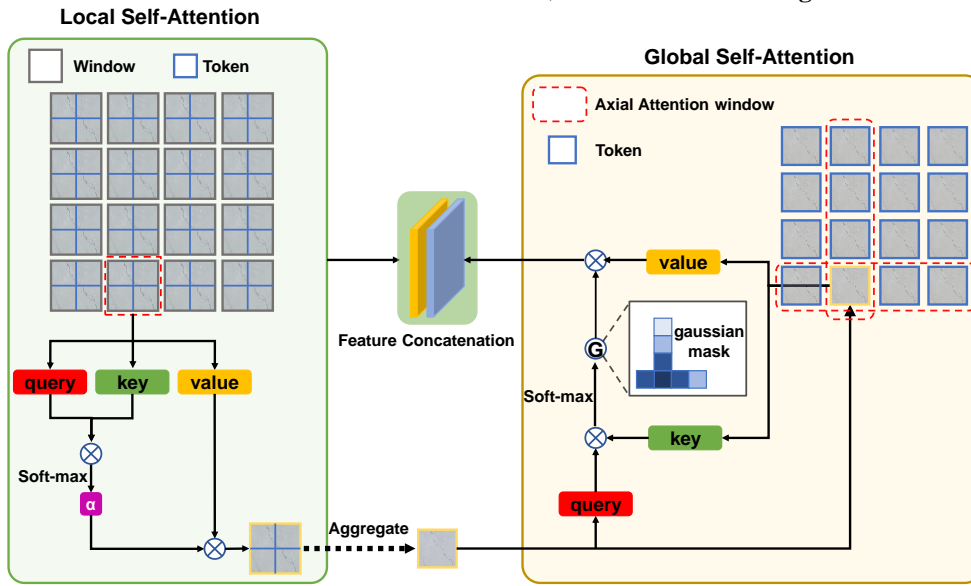

Fig. 7. Detailed architecture of LGG-SA.

## III. Datasets, Loss Functions, and Metrics

### A. Datasets

In order to apply these models in the surface cracks detection of the pavement, we develop a dataset focusing on concrete pavement segmentation. The photos of concrete pavement in this dataset were taken from the road on the Qianfoshan campus of Shandong University and collected from multiple angles under different circumstances. These datasets contain pictures with elongated cracks, grid-like cracks, etc., and those with noises, such as shadows, oil pollution, and pavement facilities.

The original size of the crack image in the dataset is 4592 × 3448. After cropping and compression, a total of 711 RGB images with a size of 224 × 224 pixels are generated. 80% of the data is used for training, and the rest is used for verification. The cracks in the 711 images were manually segmented by the LabelMe program and converted into binary images. **Fig. 8** illustrates several original crack images and their crack labels. The top panel shows the RGB images with single or multiple surface cracks as well as the ones with some noise such as shadows of leaves. And the bottom panel is ground truth with labels marked by LabelMe.

### B. Loss function

In our crack segmentation experiments, there are large background regions and uneven positive and negative samples. Therefore, we use a variety of loss functions commonly used in image segmentation in the training process and select the most suitable loss function by applying them to the MTUNet network.

*1) Binary entropy loss (BCE loss):* BCE loss is a pixel-level cross-entropy loss. In other words, the cross-entropy loss is calculated pixel by pixel for the predicted image and averaged on all pixels. We propose that the predicted output after sigmoid transformation is $y_{pred}$, then BCE loss can be written as:

$$l_{BCE} = -[y_{true} \cdot \log y_{pred} + (1-y_{true}) \cdot \log(1-y_{pred})] \quad (13)$$

*2) Dice loss:* The dice loss function is proposed by Milletari et al. [35] to deal with the strong imbalance of positive and negative samples in semantic segmentation. We define the dice coefficient as follows:

$$Dice = \frac{2|X \cap Y|}{|X|+|Y|} \quad (14)$$

where X and Y represent the segmentation Ground True and the prediction result, respectively. From this, the formula of dice loss is obtained by

$$l_{Dice} = 1 - \frac{2|X \cap Y|}{|X|+|Y|} \quad (15)$$

*3) Linear combination of BCE loss and dice loss:* In this part, the loss function adopts a linear combination of the BCE loss function and dice loss function, where the weights of the former and the latter are adjustable. The loss functions during our training can finally be written as (16) and (17).

$$l_{combine1} = 0.5 \cdot l_{BCE} + l_{Dice} \quad (16)$$

$$l_{combine2} = l_{BCE} + l_{Dice} \quad (17)$$

*4) Lovasz loss:* Lovasz loss is a loss function proposed by Maxim Berman et al. [36]. It directly optimizes IOU and is based on a convex Lovasz extension of submodular losses. It performs better than traditional BCE loss on image segmentation tasks.

### C. Model evaluation metrics

There are various metrics to measure model performance in deep learning. We applied the indicators, including Intersection over Union (IoU), Accuracy, Precision, Recall, and F1 score, to comprehensively measure the various crack segmentation models in this paper. The details are described below. **Fig. 9** shows the simple expression of crack prediction results and ground truth.

Intersection over Union (IoU) frequently serves as a performance metric in semantic segmentation and object detection. In the second-class segmentation, it can be written in the following

$$IoU = \frac{TP}{TP+FP+FN} \quad (16)$$

Accuracy is the most basic and most commonly used indicator in deep learning. In semantic segmentation, it represents the proportion of correctly predicted pixels out of all pixels. The formula can be written as

$$Accuracy = \frac{TP+TN}{TP+FP+FN+TN} \quad (17)$$

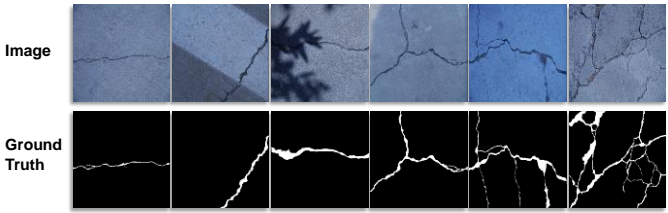

**Fig. 8.** Dataset images of raw RGB images and binary labels.

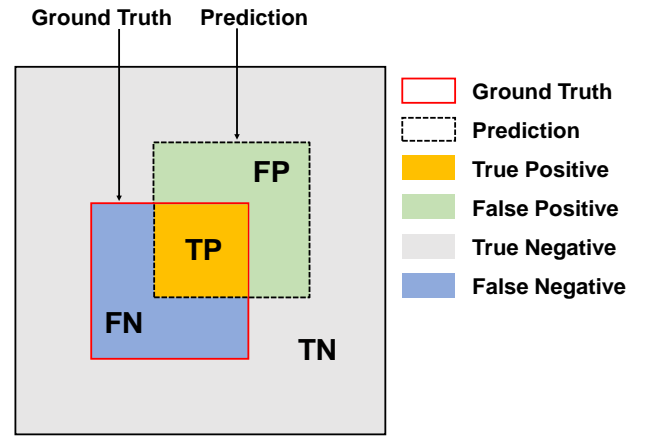

**Fig. 9** Schematic of crack prediction results and ground truth.

In addition, Precision and Recall are also commonly used indicators in the field of deep learning, and they are often used together, as shown in (18) and (19).

$$Precision = \frac{TP}{TP+FP} \quad (18)$$

$$Recall = \frac{TP}{TP+FN} \quad (19)$$

It is worth noting that when one of the indicators of Precision and Recall is high, it is difficult for the other indicator to be very high. Therefore, the F1 score is introduced to seek the best balance between Precision and Recall. The F1 score can be written as

$$F1 = 2 \times \frac{Precision \times Recall}{Precision + Recall} = \frac{2 \times TP}{2 \times TP + FP + FN} \quad (20)$$

## IV. RESULTS AND DISCUSSIONS

### A. Model training

This section presents specific details in model training, including optimizer selection, hardware facilities, and software facilities. All networks are based on the PyTorch [37] work package, and these networks were performed on hardware (CPU: Intel® Xeon® Gold 6248R CPU @3.00GHz, RAM:256GB, GPU: NVIDIA Geforce RTX 3080). In the experiment, each model uses the Adam [38] algorithm to optimize the hyperparameters, which has been proven to be the most efficient and fastest way to reduce bias. In the parameter selection, the initial learning rate is set to 0.001and the weighted delay is 0.0001. The eps is set as 10-8 in case of a zero denominator, and the first and second-moment estimates are 0.9 and 0.999, respectively. For the performance evaluation of different models, the batch size of each model is set to 12, and the number of epochs is set to 1000.

In order to select the efficient loss function for different deep learning models and make a fair comparison, we first test the loss functions of Lovasz, BCE, Dice, and their combination by the MTUNet model. As shown in **Fig.10**, the loss of BCE converges fastest among all lost functions, and the Dice function is inferior to the BCE. The loss of Lovasz converges slower than any other one. On the other hand, we evaluate the F1 and IoU performance metrics of the MTUNet Model with different loss functions. Lovasz presents better performance than all other ones except for the combination of 0.5BCE and Dice loss functions. Considering the convergence and performance, we choose the combination of 0.5BCE and Dice loss function to make a comparison among all different models.

In this study, we adopted and trained both transformer-based and CNN-based crack segmentation models to show the superiority of the SA algorithm in crack identification. The former includes TranUNet, SwinUNet, and MTUNet, which are recently developed for semantic segmentation. CNN-based models are widely used in semantic segmentation and perform well in various segmentation tasks. Here, we adopted, modified, and trained five typical CNN-based models based on our own dataset, that is DeepLabv3+, UNet, UNet++, ResUNet, ResUNet++,AttnUNet. DeepLabv3+ uses atrous convolution and atrous spatial pyramid pooling; UNet++ integrates UNet of different depths on the original decoder-encoder structure; ResUNet combines residual network and UNet; ResUNet ++ introduces squeeze-and-excitation blocks and atrous spatial pyramid pooling on the basis of ResUNet; AttnUNet is based on UNet and added an Attention Gate module to make an attention mechanism for skip connection and up-sampling layers.

**Fig. 11** shows the loss of 0.5BCE+Dice of nine semantic models during the training process within 1000 epochs. The loss of all those models converges within 1000 epochs. Among those models, SwinUNet converges much faster than any other one. When the epoch is greater than 800, the MTUNet has the smallest loss. Both MTUNet and TransUNet converge slower than the SwinNet at an early stage during the training, but their losses are lower than that of TransUNet at a late stage. DeepLabV3+ converges the slowest and has the largest loss during the training process. As a result, we can further compare their performance based on the validation and test data between these semantic models with the 0.5BCE and Dice loss.

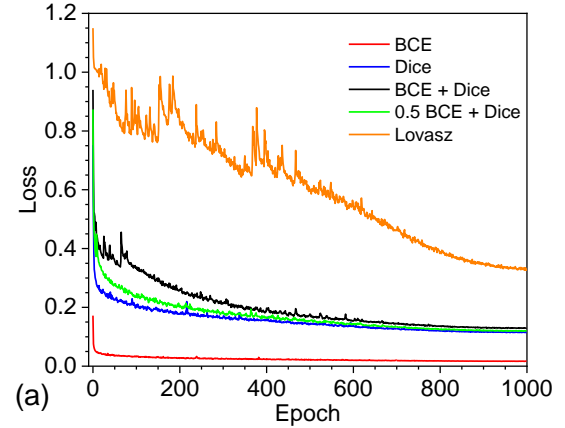

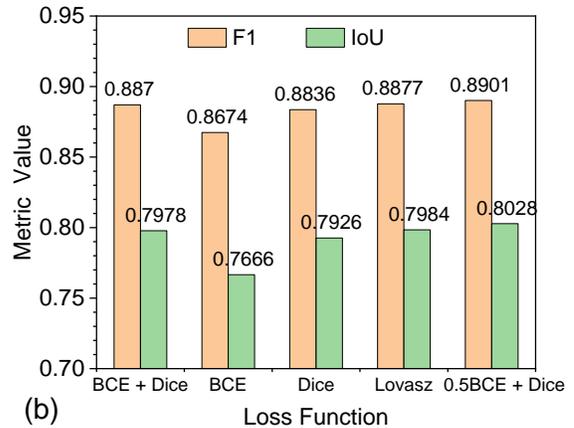

Fig.10 (a) The loss function values of the MTUNet model during the training process. (b) The F1 and IoU metric values of MTUNet for different loss functions.

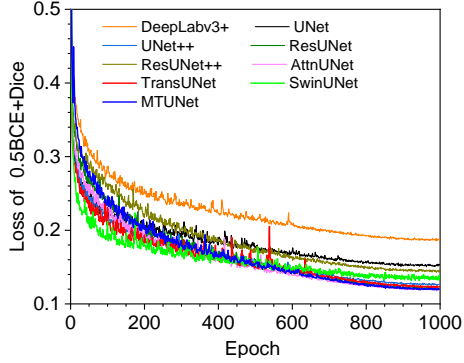

Fig.11. The loss of 0.5BET+Dice of semantic models.

*B. Evaluation of models by metrics and performance on the validation dataset*

Table I lists the evaluation metrics of these models on the same validation set after training on the same training set by taking values from the first fold. Here, all the hyperparameters of semantic models are not trained in advance except for the SwinUNet which uses weights on medical image segmentation as pre-trained parameters. The reason is that SA has no obvious effect on low-level feature learning. We used the transfer learning strategy to set the pre-training weights for the pure transformer model of SwinUNet. In addition, the speed of image processing is averaged based on the data in the test dataset. We assume that the speed of the image processing is the same since we can not simply calculate the value in the validation procedure. The unit of img/s indicates the number of images that can be calculated per second, the training loss represents the loss function value when training reaches the 1000th epoch, and the metrics are averaged over the last 100 epochs during training.

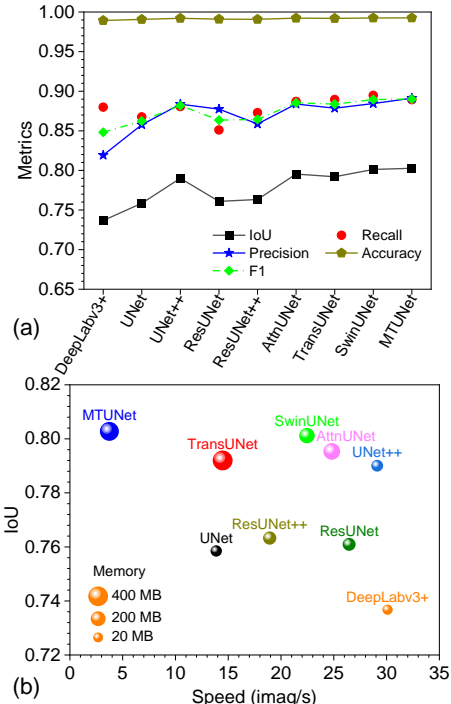

Fig. 12. (a) Metrics and (b) performance of validation set for various semantic models.

TABLE I
PERFORMANCE DETAILS AND METRICS OF DIFFERENT MODELS ON THE VALIDATION SET.

| Models | Speed (img/s) | Memory (MB) | Training loss | IoU | Recall | Precision | Accuracy | F1 score |
|---|---|---|---|---|---|---|---|---|
| DeepLabv3+ | 30.1 | 22.4 | 0.1867 | 73.7% | 88.0% | 81.9% | 98.9% | 84.8% |
| UNet | 13.9 | 30.0 | 0.1530 | 75.9% | 86.8% | 85.8% | 99.1% | 86.2% |
| UNet++ | 29.1 | 35.0 | 0.1258 | 79.0% | 88.1% | 88.4% | 99.2% | 88.2% |
| ResUNet | 26.5 | 49.8 | 0.1200 | 76.1% | 85.1% | 87.7% | 99.1% | 86.4% |
| ResUNet++ | 18.9 | 55.3 | 0.1445 | 76.3% | 87.3% | 85.9% | 99.1% | 86.4% |
| AttnUNet | 24.8 | 153.0 | 0.1192 | 79.5% | 88.7% | 88.4% | 99.2% | 88.5% |
| TransUNet | 14.5 | 401.0 | 0.1225 | 79.2% | 89.0% | 87.9% | 99.2% | 88.4% |
| SwinUNet | 22.5 | 105.0 | 0.1340 | 80.1% | 89.5% | 88.4% | 99.2% | 88.9% |
| MTUNet | 3.7 | 301.0 | 0.1199 | 80.3% | 88.9% | 89.1% | 99.3% | 89.0% |

**Fig. 12** and Table I indicate that transformer-based models have obvious advantages, such as accuracy, recall, precision, IoU, and F1 score, over the CNN-based models based on the results of the validation set. Accuracy provides an evaluation of the entire dataset and is a powerful metric for the class-balanced dataset. **Fig.12a** shows that all models have very high accuracy over the 98.9% in the whole image domain, and transformer-based models are slightly higher than the CNN-based models. The reason is that crack pixels occupy a small portion of the entire image and are class-unbalanced, and all those models perform great in the intact region. Recall and precision are two competing metrics and emphasize different aspects of positive prediction. Recall indicates the capability of the model to recognize the cracks. The bigger the prediction region, the higher probability of the true positive region. The recalls of the MTUNet and SwinUNet algorithms have reached 89.0% and are higher than those of CNN models. Among CNN models, DeepLabv3+, UNet++, and AttnUNet have scores of recall above 88.0% and are higher than other models. Precision describes how accurate the predicted cracks are. The smaller the prediction region, the higher precision the model may have. UNet++ and AttnUNet have a precision of 88.4%, which is higher than any other CNN models. The precisions of SwinUNet and MTUNet are higher than 88.4%, while that of TransUNet is slightly lower than 88.4%. IoU and F1 scores are two typical evaluation metrics for the unbalanced dataset, which combine precision and recall. Usually maximizing the IoU and F1 scores implies improving the precision and recall simultaneously. The F1 score is the harmonic mean of precision and recall. IoU puts a lower weight on the TP in the equation. Compare to the F1 score, IoU usually is lower than the F1 score and varies mildly with the TP. As shown in **Fig.12 a**, the F1 score and F1 show the same trend. The IoU and F1 scores of SwinUNet and MTUNet are higher than the CNN models, and the scores of TransUNet are higher than the CNN models except for AttnUNet. In a word, the scores of evaluation metrics of the transformer-based models are higher than those of CNN models except that TransUNet has lower scores than AttnUNet by several metrics.

**Fig.12 b** presents the efficiency and memory consumption along with the IoU averaged over the last 100 epochs listed in Table I for all models. This experiment records the speed of image processing in the testing session and the memory usage of the model. The comprehensive performance of the model in **Fig.12 b** describes the model memory size by the diameter of the bubble, the abscissa represents the processing speed during testing, and the ordinate represents the model IoU score. In this figure, the optimal bubble should lie in the right top corner in the smallest size. In the figure, it is obvious that the models of high IoU scores are usually large in diameter, which means they are memory-consuming. Those models include MTUNet, TransUNet, SwinUNet, and AttnUNet. Among them, MTUNet and TransUNet models have a low speed of image processing, because the model design is more complicated to effectively combine the CNN and the transformer. They all lag behind other models in computational speed and require more built-in parameters. The SwinUNet model has advantages in many aspects such as model size, model accuracy, and operation speed owing to its simple decoder-encoder architecture and the idea of sliding windows. It is worth noting that since SA has no obvious effect on low-level feature learning, we use the transfer learning strategy to set the pre-training weights for the pure transformer model of SwinUNet. Interestingly, DeepLabv3+ has the highest speed of image processing and the smallest memory usage among all models, which is 30.1 img/s and 22.4 MB, although its IoU score is the lowest in these models.

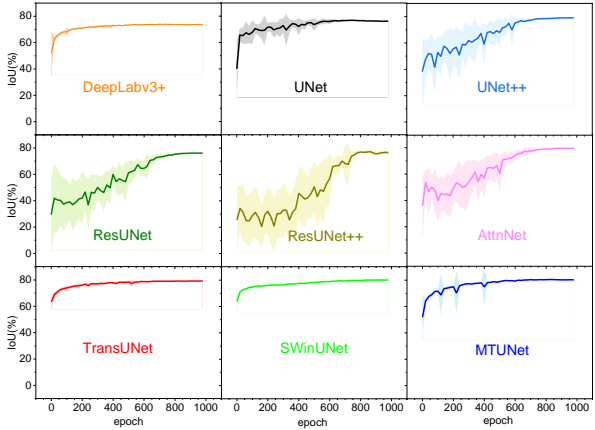

Fig. 13. The IoU curve of each model of the first fold

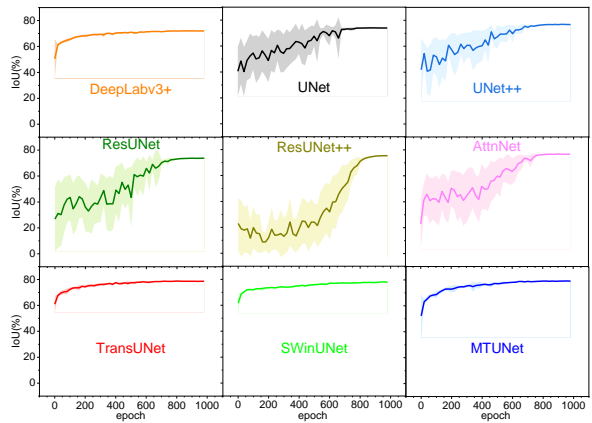

Fig. 14. The IoU curve of each model of the second fold

The convergence of the model can be further evaluated by the IoU of the validation dataset during the training process as shown in **Fig. 13**. Each IoU score represents the mean of 20 epochs during the training process, and the error bar stands for the standard deviation of these data. It can be seen that when the training time is insufficient and the number of iterations is small, the model IoU index presents a big fluctuation around the mean value and a big shadow area in the figure, which means it is difficult to achieve a stable convergence value on the validation set. We find that the traditional UNet segmentation network based on CNN has a serious overfitting phenomenon, indicated by a big shadow area in the figure. In addition to UNet, we also found the same phenomenon on other improved networks based on UNet, such as UNet++, ResUNet, AttnUNet, and ResUNet++. In other words, when training on a small scale, these traditional CNN-based segmentation models are difficult to perform stably during validation, and it is difficult for these models to achieve desirable results when changing the validation set or adding other disturbances. Interestingly, DeepLabv3+ exhibits a small fluctuation around its mean of the IoU and a small shadow area, compared with other CNN models. On the other hand, transformer-based models have extraordinarily low fluctuation around their mean values and converge very quickly during the training. There are several severe fluctuations for the MTUNet model at several spots due to the noise of random allocation of the dataset.

To verify the experimental results in **Fig11**, inspired by k-fold cross-validation, we also draw the IoU curve of the model when the training and validation sets are replaced, as shown in **Fig. 14**. We can drive the same conclusion as the results of the validation dataset of the first fold. And the fluctuation is much smaller compared with the result of the first fold shown in **Fig. 13.** In summy, the convergence of transformer-based models represented by TransUNet and SwinUNet is much better than the CNN-based models.

*C. Results of crack segmentation by the test dataset*

**Fig.15** depicts the IoU, F1 score, and IoU histogram of each model on the test dataset. The test dataset contains 140 images, which were chosen randomly from the data apart from the validation and training data. The trend of IoU and F1 scores for the test dataset are the same as those of the validation dataset. The values of the test dataset are slightly higher than those of the validation dataset for each model but ResUNet++ (see **Fig.15 a and b**). The IoU and F1 scores of the ResUNet++ are improved by 2.33% and 1.56%, respectively. Those for all other models are less than 0.98% and 0.64%, respectively. The detailed values of all models are listed in Table 2. The reason is that the hyperparameters of the model for the test dataset are well-trained, which demonstrates that the models are reliable for surface crack `segmentation. The obvious improvement for the test dataset of ResUNet is due to the significant improvement of a portion of data as shown in **Fig 15.e**. This also shows that the histogram of the metrics may provide more information to evaluate the models.

**Fig. 15 c~l** shows the metrics variation of each model by providing the IoU histogram distribution of each model. A better model should exhibit a higher mean value of IoU and a narrower bandwidth. Shown in **Fig.15** by (d~l), histograms of IoU are plotted for each model, the Gauss function is chosen for the nonlinear curve fitting to describe the distribution of IoU, and the mean and full width at half maxima (FWHM) along with standard errors are summarized in **Fig.15 c.** On the test dataset, the mean value of IoU increases from DeepLabv3+ to MTUNet, and transformer-based models have higher mean values of IoU than CNN-based models in IoU generally. Consistent with the finding from the validation dataset, the IoU of DeepLabv3+ is the smallest among all models while the AttnUNet is the highest in all CNN-based models. The IoU value increases from TransUNet to SwinUNet, MTUNet. The value of TransUNet and SwinUNet is lower than that of AttnUNet. On the other hand, the FWHM decreases from DeepLabv3+ to MTUNet and transformer-based models have a narrower distribution of IoU around the mean value than CNN-based models in general. The FWHM values of transformer-based models are lower than all CNN-based models.

The values of DeepLabv3+ and AttnUNet define the upper and lower bounds of the CNN-based models in this study. SwinUNet has the lowest FWHM and TransUNet has the highest value in transformer-based models. In summary, the transformer-based models have a better performance in IoU than CNN-based models. TransUNet is the best transformer-based model and TransUNet is an inferior transformer model among these three models. For the CNN-based models, AttnUNet is preferred and DeepLabv3+ is the worse candidate in respect of the IoU score.

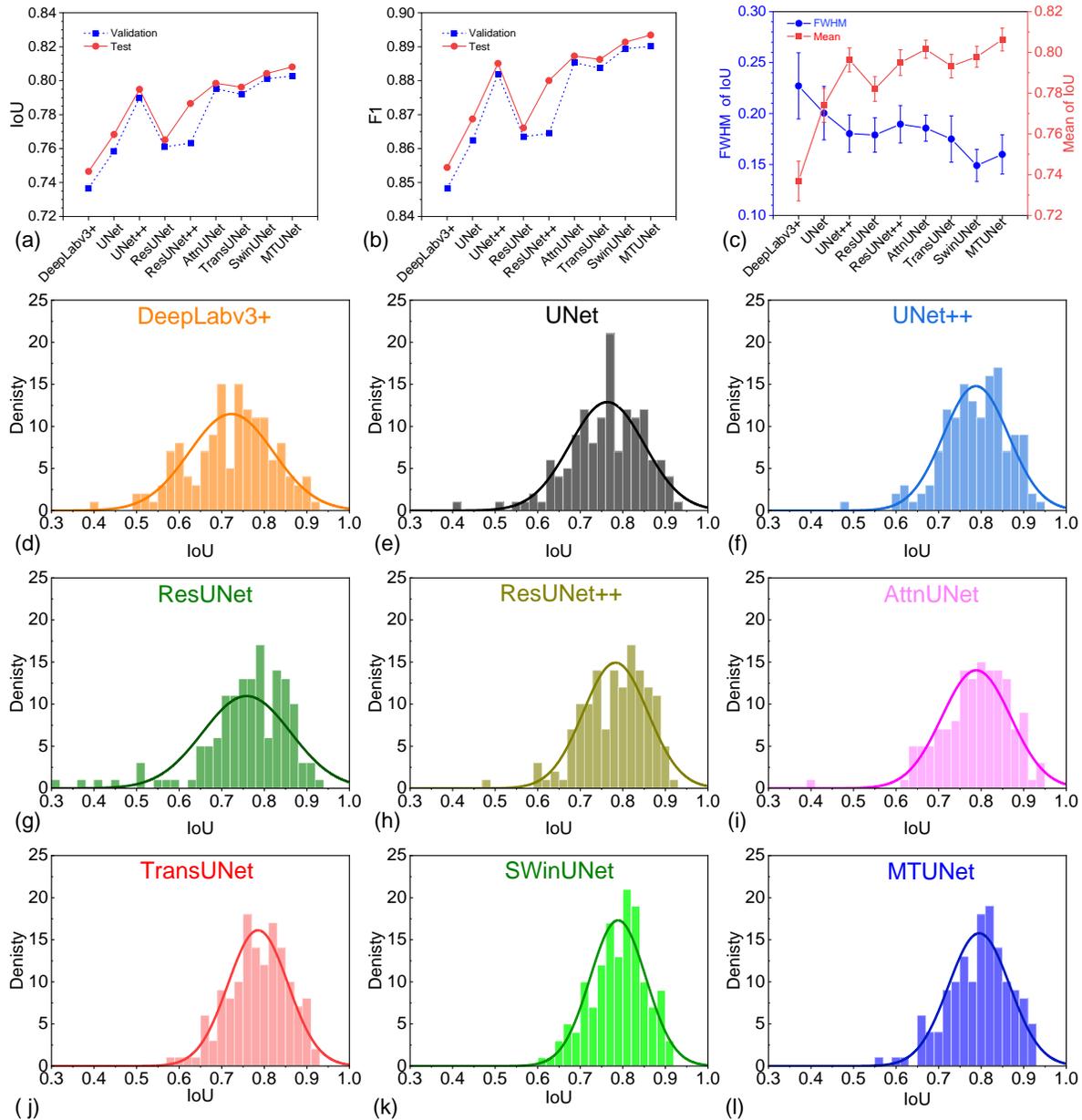

Fig. 15. Evaluation metrics and histogram of each segmentation model on a test dataset.

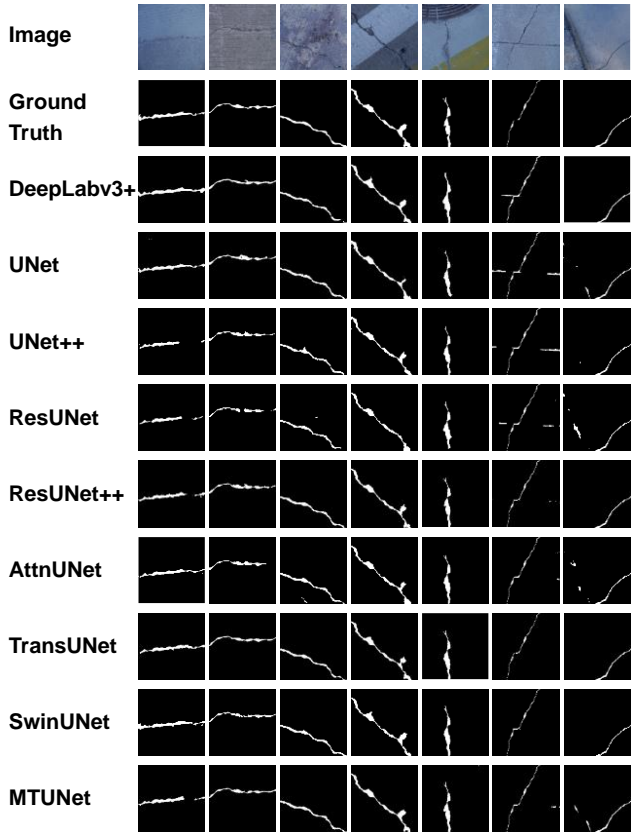

Fig. 16 Segmentation of single cracks by all models.

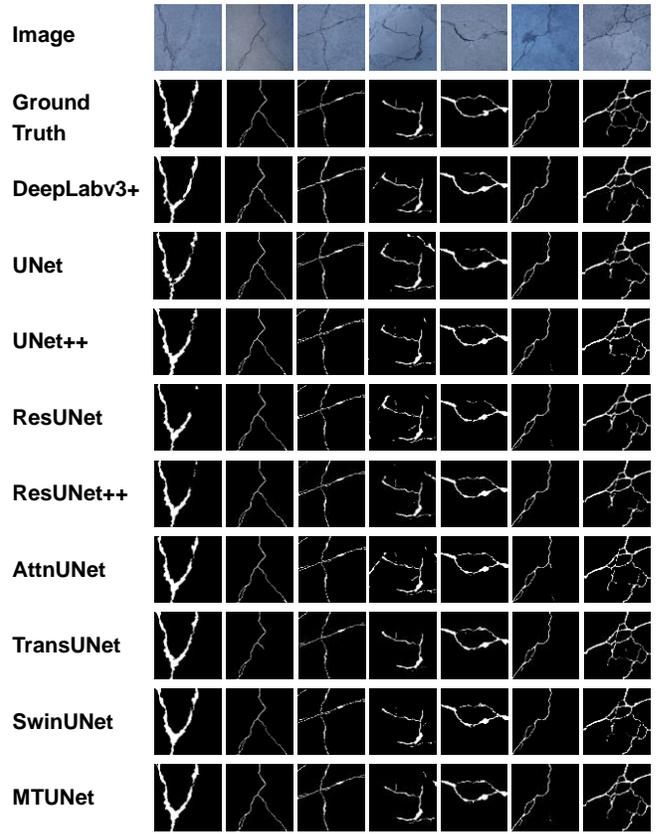

Fig. 17 Segmentation of complicated cracks all by all models.

TABLE II
IoU AND F1 SCORES OF ALL MODELS WITH OPTIMAL PARAMETERS ON THE TEST DATASET

| Models | DeepLabv3+ | UNet | UNet++ | ResUNet | ResUNet++ | AttnUNet | TransUNet | SwinUNet | MTUNet |
|---|---|---|---|---|---|---|---|---|---|
| IoU | 74.65% | 76.83% | 79.49% | 76.51% | 78.66% | 79.85% | 79.62% | 80.43% | 80.81% |
| F1 score | 85.44% | 86.87% | 88.51% | 86.61% | 88.00% | 88.73% | 88.63% | 89.14% | 89.35% |

In order to show the experimental results more intuitively, we established the test set of 140 images to visualize the segmentation results of each model, as shown in **Fig.16** and **Fig.17**. The former illustrates the results of the seven images with a single crack, and the latter are those with interconnected cracks to illustrate the performance of complicated cases. Each column in these two figures represents one case and illustrates the image, label, and outputs by all models. As shown in **Fig.16**, the first column shows the results for a single central crack in the image. Considering the completeness of the crack, a part of the crack is missing in the results by UNet++, ResUNet, ResUNet++, and MTUNet models. In the second case, a small part of the crack is missing in the outputs of ResUNet and AttnUNet. We can find that each model can successfully get rid of the noise introduced by the shadow and painting from the third to the fifth cases. In the last two cases, environmental noises caused by the construction are introduced to show the reliability of the models. The identified cracks are contaminated by some segment of the concrete block boundary or spots in DeepLabv3+, UNet, UNet++, ResUNet, AttnUNet, and MTUNet models. But the spots in the cases of AttnUNet and MTUNet are discrete and can be removed simply by considering the continuity of the shapes of the cracks further. ResUNet++, TransUNet and SwinUNet show great potential in getting rid of the noises.

**Fig.17** illustrates the cases with branched cracks and other complicated cracks. For the branched cracks depicted by the first three columns, ResUNet, ResUNet++, and TransUNet missed a portion of the cracks. The fourth case shows the influence of the noise due to operation. DeepLabv3+, UNet, UNet++, ResUNet, ResUNet++, AttnUNet, and MTUNet introduce some noises and TransUNet misses a small portion of crack, which is very thin. The fifth and sixth cases show images with an area enclosed by cracks. DeepLabv3+, UNet, and ResUNet can not identify the enclosing crack entirely and miss some parts. The spot noises in the sixth case can not be removed by UNet++, ResUNet, and AttnUNet. The last case shows the image with very complicated cracks with different levels of thickness, and all models miss a part of the cracks of the thinnest level. In conclusion, all models can identify most of the cracks and can get rid of noise due to construction and operation such as shadows, oil painting, and block boundaries to some degree. In general, transformer-based models are superior to CNN-based models. Among three transformer-based models, SwinUNet is the optimal model. However, there is still some space to further improve the model to identify the entire cracks for complicated cases with different thickness levels.

*D. Prospect analysis*

The boom in deep learning of the early years was driven by the multilayer perceptron, convolutional network, and recurrent network architectures. Models of classical convolutional neural networks dominate the tasks of computer vision and surface crack identification region. Recently, transformer architecture has shown up as a promising model for various computer vision tasks. Compared with the mature CNN, it has a broader development space. Here, we have demonstrated that a crack detection model using transformers can achieve or even exceed the performance of previous CNN models. Transformer architecture-based crack detection model may become the main way to realize crack detection through deep learning in the future.

However, combined with the research in this article, the transformer model needs to be improved in terms of model memory and computational efficiency. The encoder-decoder architecture that combines transformers and CNNs makes the model more complex, slower to compute and requires better GPUs for training. Considering the practical application cost, a modest improvement in the model is still required. Moreover, there is still space to develop models for the task with complicated cracks of different levels of thickness.

## V. Conclusion

In order to explore surface crack identification by artificial intelligence, we created a dataset of images of pavement for a total of 711 images and investigated nine deep-learning models to evaluate their performance in surface crack detection. For nine models, DeepLabv3+, UNet, UNet++[20][20][20], ResUNet, ResUNet++, and AttnUNet are typical CNN-based models used in computer vision, and the rest models of TransUNet, SwinUNet, and MTUNet are emerging transformer architecture. We considered surface cracks including single cracks and complicated cases and interference such as shadows, oil pollution, and regular pavement joints by taking pictures of the on-campus road in service to reflect their performance in multi-environment in practical application in the future.

Sevel loss functions including Lovasz, BCE, Dice, and their combination with various proportions of Dice and BCE by the MTUNet model. The model can be easily trained with the BCE loss function with a few epochs. From the perspective of the difficulty in training the model, BCE has the highest priority and is followed by Dice and their combination, and the Lovasz is the worst to train the hyperparameters. However, the preference is changed when the metrics of F1 and IoU are evaluated. The loss of 0.5 BCE combined with Dice is better than that of Lovasz, BCE+Dice, Dice, BCE in sequence. As a result, the 0.5BCE+Dice is preferred as the loss function for the models.

All the models are evaluated by a variety of accuracy indicators to evaluate the level of various crack segmentation models, including IoU, recall, precision, accuracy, and F1 score. The values of the last 100 epochs in the iterative process were averaged to reduce the accidental error, each model evaluation index selects. The results of the validation set show that transformer-based models generally have obvious advantages over CNN-based models in accuracy, recall, precision, IoU, and F1 score. However, transformer-based models usually require larger memory and a relatively lower speed of image processing. Among CNN-based models, UNet++ and AttnUNet have larger metrics of IoU and F1 scores and a larger speed of image processing. Among three transformer-based models, the IoU and F1 scores increase from TransUNet to SwinUNet and MTUNet, the processing efficacy decreases from SwinUNet to TransUnet and MTUNet, and the memory consumption rises from SwinUNet to MTUNet and TransUNet. Interestingly, DeepLabv3++ has the lowest values of IoU and F1 but the highest processing efficiency and smallest memory requirement.

The accurate performance of nine models was further evaluated by the fluctuation of the IoU curves during the training process. The transformer-based models present lower fluctuation while CNN-based models show extremely larger fluctuation around its mean value of every 20 epochs except for DeepLabv3+, especially at their early training stages. Among transformer-based models, SwinUNet and TransUNet are better than MTUNet for a smaller fluctuation. The encoder-decoder architecture based entirely on CNN is unstable when the training time is insufficient. At this time, the transformer model shows better stability.

The results of the test dataset exhibit the same trend as the validation dataset. The histogram displays that transformer-based models have high metric values and smaller full width at half maxima values. Finally, the segmentation results of each model illustrate their performance intuitively by input figures with a single crack, branched cracks, complicated cracks, and road disturbances like an oil painting. The merits of transformer-based models were further confirmed directly. The defects that can be identified directly include the missing thin crack branch and false judgment of building block boundaries. In all, this research shows that the transformer-based crack segmentation model is better than the CNN model, and the SwinUNet outperforms all others. But at the same time, the computational complexity of the transformer model is higher, the operation speed is slower, and there is still room for improvement in the model structure.

## Acknowledgment

We gratefully acknowledge the financial support of the Shandong Provincial Natural Science Foundation (No. ZR2021MA045).